# Automated speech audiometry: Can it work using open-source pre-trained Kaldi-NL automatic speech recognition?

Gloria Araiza-Illan[1,2], Luke Meyer[1,2], Khiet P. Truong[3] and Deniz Başkent[1,2]

[1] Department of Otorhinolaryngology, Head and Neck Surgery, University Medical Center Groningen, University of Groningen, Groningen, the Netherlands

[2] W.J. Kolff Institute for Biomedical Engineering and Materials Science, University Medical Center Groningen, University of Groningen, Groningen, the Netherlands

[3] Human Media Interaction, University of Twente, Enschede, the Netherlands

Corresponding author:

Gloria Araiza-Illan

e-mail: g.a.araiza.illan@rug.nl

**Abstract**

A practical speech audiometry tool is the digits-in-noise (DIN) test for hearing screening of populations of varying ages and hearing status. The test is usually conducted by a human supervisor (e.g., clinician), who scores the responses spoken by the listener, or online, where a software scores the responses entered by the listener. The test has 24 digit-triplets presented in an adaptive staircase procedure, resulting in a speech reception threshold (SRT). We propose an alternative automated DIN test setup that can evaluate spoken responses whilst conducted without a human supervisor, using the open-source automatic speech recognition toolkit, Kaldi-NL. Thirty self-reported normal-hearing Dutch adults (19-64 years) completed one DIN+Kaldi-NL test. Their spoken responses were recorded, and used for evaluating the transcript of decoded responses by Kaldi-NL. Study 1 evaluated the Kaldi-NL performance through its word error rate (WER), percentage of summed decoding errors regarding only digits found in the transcript compared to the total number of digits present in the spoken responses. Average WER across participants was 5.0% (range 0 - 48%, SD = 8.8%), with average decoding errors in three triplets per participant. Study 2 analysed the effect that triplets with decoding errors from Kaldi-NL had on the DIN test output (SRT), using bootstrapping simulations. Previous research indicated 0.70 dB as the typical within-subject SRT variability for normal-hearing adults. Study 2 showed that up to four triplets with decoding errors produce SRT variations within this range, suggesting that our proposed setup could be feasible for clinical applications.

**Introduction**

For around 80 years, speech audiometry has been considered a basic instrument for audiological evaluation, used for multiple purposes from research to rehabilitation (Boothroyd, 1968; Jerger et al., 1968). The digits-in-noise (DIN) test was proposed as a practical and reliable speech audiometry test, as it is minimally dependent on linguistic skills and feasible for a wide range of listeners, including children and adults with different hearing statuses (Smits et al., 2013). In the Netherlands, the predecessor of the DIN test, the Digit Triplet Test (DTT), was introduced as a simple speech-in-noise test that would be easy to implement and to run in the clinic as part of speech audiometry, or by the listener themselves via online platforms or on the telephone. Perhaps due to these advantages, the DTT became the National Dutch Hearing test, and was widely used for identifying difficulties regarding speech understanding in noise [for a review, see Van den Borre et al. (2021)]. The test has been shown in multiple studies to have both high test-retest reliability and a correlation with average pure-tone thresholds for both normal-hearing and hearing-impaired adults. In addition, it has been implemented in different languages (Zokoll et al., 2012) and on different interfaces (such as telephone, online, and tablets, since it only requires input via a keypad). It has been used to screen various populations, from young children to older adults, both normal hearing and hearing-impaired (Van den Borre et al., 2021). Ten years after the introduction of the DTT, the DIN test was presented. One of the main differences between the two tests is that the DIN test uses a larger bandwidth, with an upper limit of 16000 Hz, compared to the 300–3400 Hz bandwidth of the DTT (Smits et al., 2013). The DIN test has been compared to the standard Dutch speech audiometry speech-in-noise test (Plomp & Mimpen, 1979), which uses sentences as speech material and has been the gold standard for the assessment of speech recognition in noise abilities, showing a correlation coefficient of 0.96 (Smits et al., 2013). The reported within-subject variability for young normal-hearing individuals (21 to 32 years old), calculated from the standard deviation of 25 SRT estimates, was 0.70 dB (Smits et al., 2013).

The DIN test uses recorded digits, presented in triplets. It is assumed that using the digits in a closed-set paradigm requires very little linguistic or top-down processing. This assumption is based on the data from the Spoken Dutch Corpus project (Oostdijk, 2000), where the digits were in the top 500 most frequently spoken words in Dutch, are known by children from a young age, and tend to be among the first words learned as a second language. The test is widely used with either a clinical or an online setup. The clinical setup requires a human supervisor in charge of controlling the presentation of each triplet and scoring the listener's spoken responses, limiting the test to clinical settings and the availability of qualified personnel to run and score the test. The online setup requires that the listener familiarises themselves with the keypad of the electronic device on which the test is run and automatically scored, such as a tablet or laptop, to manually enter the digits they have heard. This setup could be challenging for some individuals, because it entails processing information both auditorily and visually, which, according to the cognitive theory of multimedia learning (Mayer & Moreno, 1998), might lead to a potential cognitive overload.

The main advantage of the clinical setup is the involvement of one information processing channel only (auditory), while the main advantage of the online setup is less reliance on a human supervisor and a reduced burden on clinicians. Here, we propose a new DIN test setup consisting of a laptop presenting the triplets and automatically scoring the participant's spoken responses through the open-source automatic speech recognition (ASR) system, readily available for Dutch, Kaldi-NL (Stichting Open Spraaktechnologie, 2022). The proposed configuration potentially improves DIN test accessibility, by offering an easy implementation and opportunity for widespread use.

*Speech Audiometry and Automatic Speech Recognition Systems*

Automatic speech recognition has been an important research area for decades, gaining more attention through its potential for a wide range of applications, such as telephony (i.e., mobile devices), military, customer service, virtual speech assistants, gaming and home automation (Cutajar et al., 2013; Jamal et al., 2017). Other areas recently linked to ASR systems are hearing-aid fitting and speech audiometry. Gonçalves Braz et al. (2022) and Fontan et al. (2022) have shown that ASR can be used to assess and optimise multiple hearing-aid configurations and to determine the settings that result in maximal speech intelligibility for particular audiometric profiles. Regarding the use of ASR in speech audiometry, two approaches have been explored: one using materials from speech audiometry tests to evaluate the performance of ASR systems (Pragt et al., 2022), and another where the use of ASR systems has been proposed to automate the scoring of speech audiometry tests (Deprez et al., 2013; Meyer et al., 2015; Ooster et al., 2018, 2019, 2020; Yilmaz et al., 2014).

ASR has been used for automating adaptive speech reception threshold (SRT) measurement in two speech audiometry tests: the sentences-in-noise LIST test for Flanders and the Netherlands (Deprez et al., 2013; Yilmaz et al., 2014), and the German matrix sentence test, also known as the Oldenburg Sentence Test (OLSA), in quiet and in noise (Meyer et al., 2015; Ooster et al., 2018, 2019, 2020). Both tests are scored through the recognition of words (keywords for the LIST test) presented within sentences. For the automation of the LIST test, the ASR consisted of a two layered hidden Markov model (HMM) based recognition system (Deprez et al., 2013; Yilmaz et al., 2014). With the recordings of normal-hearing individuals, the ASR reported maximum keyword detection accuracy was 90.7% (Deprez et al., 2013). Regarding the automation of the OLSA, two studies implemented and trained the ASR toolkit Kaldi. One study tested Kaldi with recorded speech that resembled normal-hearing listeners' responses to the OLSA (Meyer et al., 2015). The other study evaluated Kaldi's performance when scoring the OLSA with normal-hearing and hearing-impaired listeners, comparing the estimated SRT through Kaldi to the SRT estimated from a human supervisor scoring the test (Ooster et al., 2018). The reported test-retest standard deviation (using both unsupervised and supervised tests) was less than that for normal-hearing and hearing-impaired individuals as reported in the literature: 0.50 dB (Brand & Kollmeier, 2002) and 0.90 dB (Wagener & Brand, 2005), respectively.

Two other studies used the smart loudspeaker Amazon Alexa to score the OLSA, one with normal-hearing listeners (Ooster et al., 2019), and the second with both normal-hearing (divided into young and older groups) and hearing-impaired listeners (divided into mild and moderate hearing-impaired) (Ooster et al., 2020). In both cases, the error rate was calculated by comparing the ASR transcript to the manually transcribed recorded responses. For the study with normal-hearing listeners only, the reported test-retest standard deviation in SRT measures was 0.71 dB, which is close to the documented 0.50 dB for normal-hearing individuals (Brand & Kollmeier, 2002). For the study with both normal-hearing and hearing-impaired listeners, young and older normal-hearing listeners showed within-subject standard deviations of 0.57 dB and 0.69 dB, respectively, which are close to the documented 0.50 dB for clinical tests (Brand & Kollmeier, 2002). Similarly, mild and moderate hearing-impaired listeners showed within-subject standard deviations of 0.93 dB and 1.09 dB, respectively, which are also close to the documented 0.90 dB for hearing-impaired individuals (Brand & Kollmeier, 2002; Ooster et al., 2020; Wagener & Brand, 2005).

The results from the aforementioned studies consistently show the robustness of these ASR systems (Deprez et al., 2013; Meyer et al., 2015; Ooster et al., 2018, 2019, 2020). However, all these studies included training of the implemented ASR systems (using particular data collection, data preprocessing, feature extraction, and implementation of different machine learning algorithms), which, added to the tests where sentences are used as speech material, increase the level of both technical and lexical complexity of the speech audiometry tests. The use of a pre-trained off-the-shelf ASR (e.g., Kaldi-NL) and simpler speech audiometry test materials, such as the digits in the DIN test, avoids such complexity, allowing easy incorporation of ASR into speech audiometry and hence providing access for a wider range of researchers, clinicians, listeners, and any other end-users.

*The Open-source ASR Toolkit Kaldi and Kaldi-NL*

Kaldi was developed in 2011, attempting to provide modern and flexible code written in C++ (ISO/IEC, 2017) that could be easily understood, modified and extended by any potential user, and offering the opportunity of creating new acoustic and language models based on specific speech corpora (i.e., with different lexical data depending on the application and language being used). This toolkit is based on finite state transducers (FSTs), implementing state-of-the-art acoustic modelling techniques, while being computationally efficient (Povey et al., 2011).

Although originally developed for English, Kaldi has been adapted for speech recognition in different languages, such as Dutch (Stichting Open Spraaktechnologie, 2022), Punjabi (Guglani & Mishra, 2018), Kannada (Yadava & Jayanna, 2017), Spanish (Becerra et al., 2016), Arabic (Bezoui et al., 2016), Russian (Kipyatkova & Karpov, 2016), Italian (Cosi, 2015), Serbian (Popović et al., 2015), and Swahili, Hausa and Amharic (Besacier et al., 2015). The Dutch version, hereinafter referred to as Kaldi-NL, was created using models based on Spoken Dutch Corpus that consists roughly of 1000 hours of adult speech (Oostdijk, 2000) by the University of Twente, with a lexicon of approximately 250000 words. Kaldi-NL showed one of the best performances on the NBest BN-NL benchmark set, which is intended to set up the infrastructure and evaluate large vocabulary speech recognition for the Dutch language (Kessens & Leeuwen, 2007), with around 10% word error rate (WER) (Ordelman & van Hessen, 2018), one of the main evaluation metrics for ASR systems.

*The Present Study*

The purpose of this study was to provide an initial evaluation of the performance of Kaldi-NL when used to decode the spoken responses of participants to automatically score the DIN test, exploring its feasibility as a fully automated speech audiometry test without requiring any extra training of the ASR. The proposed setup was tested without a human supervisor, and with normal-hearing adult participants. The recordings from the participants' spoken responses were used for offline evaluation of the automated DIN test. In Study 1, we assessed the WER of Kaldi-NL using the transcripts of the decoded responses. In Study 2, we evaluated how well the proposed automated DIN test could perform through the modelled effect of the inclusion of decoding errors from Kaldi-NL (based on Study 1) on the estimated DIN test SRT.

**Study 1: Evaluation of the ASR Kaldi-NL**

*Methods*

Our DIN+Kaldi-NL program was written in Python (Van Rossum & Drake, 1995). Firstly, the DIN test algorithm was implemented using the same parameters and structure as the clinical DIN test implementation. For one DIN test (24 triplets), each of the participant's spoken responses was decoded and transcribed using Kaldi-NL. The shell scripts (Gite, 1998) from Kaldi-NL were embedded into the main DIN+Kaldi-NL Python algorithm to automate both the DIN test and subsequent Kaldi-NL decoding. After a participant's spoken response was decoded by the ASR, the resulting transcript was processed to remove all non-digit words. The obtained digit-only transcript was then used to score the corresponding participant's response to the presented triplet.

*Participants*

Thirty native Dutch speaking adults (age range: 19–64 yr, M = 36.4, SD = 15.5) participated. All participants had self-reported normal hearing. There was no exclusion based on age or regional Dutch accent. Participant population was either Bachelor's students from the Rijksuniversiteit Groningen, or staff members from the Department of Otorhinolaryngology at the Universitair Medisch Centrum Groningen (UMCG). All participants provided written informed consent, participated on a voluntary basis, and gave permission to record their spoken responses. The study protocol was part of a larger project, approved by the UMCG Medical Ethics Committee (METc 2018/427, ABR NL66549.042.18).

*Stimuli*

For the DIN test, the Dutch speech and noise materials were taken from the 10 predefined lists in Smits et al. (2013), consisting of 24 triplets randomly chosen from a set of 120 unique triplets (each one with three different digits). The digits (0 to 9) were individually recorded by a male speaker. After a first estimation of the 50% intelligibility points and slopes of the speech recognition function, each digit was adjusted in level according to its calculated individual level corrections to obtain a maximum slope of the speech recognition function and coinciding 50% intelligibility points, for normal-hearing listeners. The digits were concatenated into triplets including intervals of 500 ± 50 ms before the first and after the last digits, and of 150 ± 50 ms between digits. The masking noise was the long-term average speech spectrum noise, and a unique noise token was assigned to each of the triplets with a duration ranging from 2.8 to 3.1 s, same duration of the triplet [for more details, see Smits et al. (2013)].

*Setup*

Both the speech and noise were diotically presented from a Lenovo ThinkPad E15 laptop at 0° azimuth in the horizontal plane through the laptop internal loudspeakers. The levels of the materials were calibrated with a Knowles Electronics Mannequin for Acoustic Research (KEMAR, GRAS, Holte, Denmark) located approximately 80 cm from the laptop loudspeakers and a Svantek sound level meter (Svan 979). Participants were seated at a desk upon which the laptop was placed. Their responses were recorded by an external microphone (Speedlink Volity Ready Streaming Starter Set SL-800010-BK) located in front of the participant next to the laptop, with a sampling rate of 40 kHz and a cardioid polar pattern. For compatibility with the ASR system, the recorded responses were down-sampled from 40 kHz to 16 kHz.

*Interface calibration*

To characterise the output of the laptop speakers, a noise signal spectrally shaped to match the averaged spectrum of the DIN stimuli was used. Replicating the experimental setup, the noise was presented at the calibrated 65 dB SPL in a sound treated room. Measurements were recorded in 1/3rd octave bands and using the same KEMAR and sound level meter setup described above (Figure 1). To compare the presentation through the laptop speakers to the original noise signal, the digitally extracted levels from the original noise signal is also included in Figure 1. The figure shows that the laptop speakers have lower 1/3rd octave bands below 630 Hz in comparison to the original noise signal which peaks at 125 and 250 Hz. As a result of calibration to the overall level of 65 dB SPL, the lack of lower frequencies likely resulted in overcompensation in the higher frequencies.

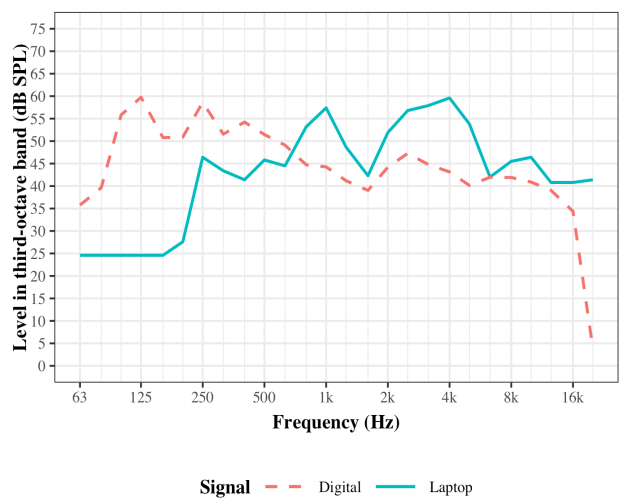

**Figure 1:** Levels in 1/3rd octave bands of spectrally shaped noise signal when presented through the laptop speakers (blue line) at the calibrated 65 dB SPL, superimposed with the digitally extracted original noise signal (red dashed line).

*Kaldi-NL*

Kaldi-NL required a working Kaldi installation, for which it is recommended, but not limited to, using a Linux computer. Kaldi-NL v0.3 was installed in the laptop and configured following the specifications listed in Stichting Open Spraaktechnologie (2022). We used the suggested acoustic and language models listed as those with best current performance (filenames NL/UTwente/HMI/AM/CGN_all/nnet3_online/tdnn, and v1.0/KrantenTT.3gpr.kn.int.arpa, respectively). The acoustic model used for Kaldi-NL corresponded to the neural network 3 setup using a time-delay neural network (TDNN). A TDNN-based setup has been suggested to perform better when compared to a regular feed-forward neural network setup (Ooster et al., 2023). The acoustic and language models used the non-telephone part of the Spoken Dutch Corpus (Instituut voor de Nederlandse Taal, 2004) as training material, which includes recordings of nearly 5000 adult speakers from all educational levels and sexes, as well as from 16 regions across the Netherlands, five Flanders regions and people outside the Netherlands and Belgium. Further details provided by L. van der Werff (personal communication, December 11, 2023) regarding the training of Kaldi-NL, include the use of the example scripts (called recipes in Kaldi's documentation) for training models using the Wall Street Journal corpus, with 40-dimensional Mel-Frequency Cepstral Coefficients (MFCCs) and additional delta, delta-delta and 100 iVector features, resulting in 220-dimensional input features.

*Procedure*

The noise presentation level was fixed at 65 dB SPL and the speech presentation level varied according to the adaptive staircase procedure within a fixed range of 42 to 75 dB SPL (the upper limit was set to prevent the adaptive test producing too high levels for presentation). The 24 presented triplets of the DIN test corresponded to one of the 10 predefined lists (Smits et al., 2013). The experiment was conducted in a quiet but otherwise not sound-treated room, to better represent clinical test rooms and include potential sound quality issues for ASR systems. The DIN test was completed within five minutes by all participants.

Algorithm 1 shows the DIN+Kaldi-NL test procedure using the same parameters and structure of the clinical DIN test implementation. At the beginning of the test, the speech level was set to 60 dB SPL (i.e., a starting signal-to-noise ratio, SNR, of -5 dB). After the first triplet was presented, if the participant's response was not evaluated to be correct by Kaldi-NL, the speech level would increase by 4 dB and the same triplet would be presented again. Only after the participant's response to this first triplet was evaluated to be correct, the speech level would decrease by 2 dB and the second triplet would be presented. A response was considered correct if all three digits with the same order as presented in the triplet were recognised as such by Kaldi-NL, after removing all non-digit words from the transcript. For all the following triplets, a step size of 2 dB was used to increase the speech level if the response was incorrect, and to decrease if correct. The outcome was given by $\mu_{SRT_{Kaldi-NL}}$ and $\sigma_{SRT_{Kaldi-NL}}$, corresponding to the mean and standard deviation, respectively, of the SNR across the triplets presented in trials 5–24 and of the last computed SNR after the 24th response. All parameters of the test and results ($\mu_{SRT_{Kaldi-NL}}$, $\sigma_{SRT_{Kaldi-NL}}$) were saved following the proposed specifications in Türüdü et al. (2022).

**Algorithm 1.** DIN+Kaldi-NL program.

```
Main input: DIN list = i
Main output: DIN test score (SRT) = (μ_{SRT_Kaldi-NL}, σ_{SRT_Kaldi-NL})
    DIN test initialisation:
        SNR_j = −5, where j=0,1
        Play triplet = list_{i,1} with SNR_0
    Presenting first triplet:
        while KaldiNL_output ≠ triplet, where KaldiNL_output = digit-only transcript do:
            SNR_1 ← SNR_1 + 4
        end while
        SNR_2 = SNR_1 − 2
    Presenting triplets 2 to 24:
        for j = 2 to 24 do:
            triplet = list_{i,j} with SNR_j
            if KaldiNL_output = triplet then:
                SNR_{j+1} = SNR_j − 2
            else:
                SNR_{j+1} = SNR_j + 2
            end if
        end for
    Calculating DIN test SRT:
        μ_{SRT_Kaldi-NL} = (Σ_{j=5}^{25} SNR_j) / 21
        σ_{SRT_Kaldi-NL} = √( (Σ_{j=5}^{25} (SNR_j − μ_{SRT_Kaldi-NL})^2) / 21 )
```

*Kaldi-NL Performance Evaluation*

The performance of Kaldi-NL was assessed offline by playing each of the recorded responses and manually annotating the digits present in the responses. The annotations were then compared to the corresponding digits transcribed by Kaldi-NL. Note that this comparison ignored non-digit words, emulating what the typical human supervisor would do when scoring the participant's responses. The WER of Kaldi-NL was calculated for each participant. WER corresponds to the percentage of summed decoding errors regarding only digits found in the transcript; i.e., insertions (addition of extra digits), deletions (omission of digits) and substitutions (replacement of digits) compared to the total number of digits present in the spoken responses. To explore potential age effects on the performance of Kaldi-NL, the population was divided into two groups, aged 19 to 41 years and 42 to 63 years.

*Results*

Figure 2 shows all individual WERs expressed in percentage. On average, 25 triplets (range 24 - 27, SD = 1) were presented in each DIN test. The average WER was 5.0% (range 0 - 48%, SD = 8.8%). Data from one participant (participant 27) seemed to be an outlier, for whom the WER was 48% (comprising 36 deletions). Excluding this participant, the WER was ≤ 13.5% for the remainder. Kaldi-NL showed 0% WER for six participants (as observed in the highlighted values in Figure 2). The WERs in Figure 2 are shown in different colours for the younger half of the group (≤ 41 years, in blue) and the older half (≥ 42 years, in orange). A group comparison did not show a significant difference, both with: $t(17.69) = 1.78, p = 0.09$; and without the outlier: $t(24.11) = 2.01, p = 0.06$.

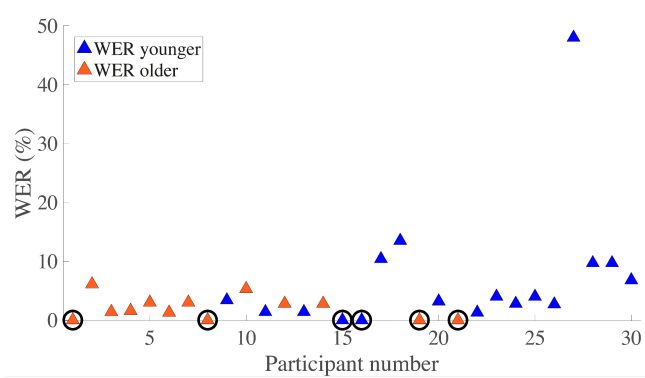

**Figure 2.** Individual WERs of Kaldi-NL for each participant, depicted as percentages. The WER for the younger group are shown in blue, whilst the values from the older group are shown in orange. The circles indicate participants for whom Kaldi-NL showed 0% WER.

Further analysis of the decoding errors showed that the errors sometimes altered the scoring of the participant's response and sometimes did not. From the total of 747 triplets presented to all participants, 80 showed decoding errors when compared to the manual annotations. As Figure 3 shows, except for the outlier (16 decoding errors), all decoding errors were in the range of 0 and 7, with an average of three triplets with decoding errors per participant. There were four cases of decoding errors by Kaldi-NL. The first two cases modified the scoring of the spoken responses: 1) where the spoken responses matched the presented triplets, but were scored as incorrect responses, and 2) where the spoken responses did not match the presented triplets, but were scored as correct responses. The other two cases did not have an impact on the scoring of the spoken responses: 3) where the spoken responses matched the presented triplets, but the transcripts included insertions of extra digits, and 4) where neither the spoken response nor the transcript (differing from the spoken response) matched the presented triplet. Table 1 shows the confusion matrix of the 80 triplets with decoding errors. The generalised probability of the effect of a decoding error affecting the score (or not) given a correct or incorrect spoken response is computed through the following equations:

$$P(e_{\text{affecting score}}|\text{correct response}) = 0.625 \quad (1)$$

$$P(e_{\text{affecting score}}|\text{incorrect response}) = 0.013 \quad (2)$$

$$P(e_{\text{not affecting score}}|\text{correct response}) = 0.125 \quad (3)$$

$$P(e_{\text{not affecting score}}|\text{incorrect response}) = 0.238 \quad (4)$$

**Table 1.** Confusion matrix of the 80 triplets for which there was a decoding error.

|  | **Correct spoken response** | **Incorrect spoken response** |
|---|---|---|
| **Automatically scored as correct** | 10 | 1 |
| **Automatically scored as incorrect** | 50 | 19 |

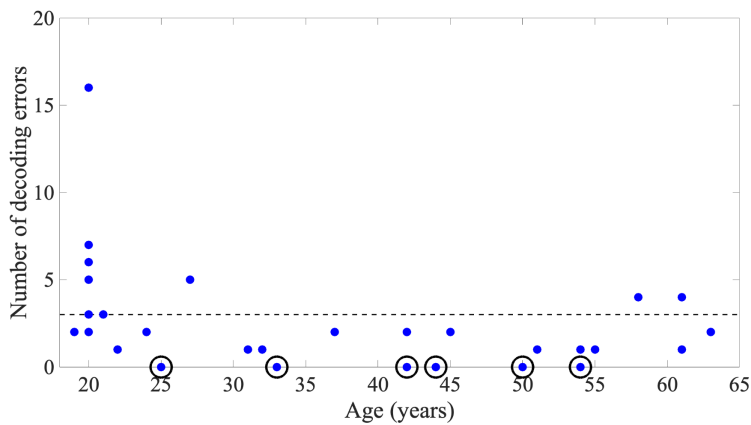

**Figure 3.** Number of triplets with decoding errors from Kaldi-NL for each participant, shown as a function of age. The dotted line corresponds to the average across all participants, namely three triplets with decoding errors. Highlighted in circles are the scores for the same six participants in Figure 2, illustrating that Kaldi-NL had no decoding errors.

All 16 insertions corresponded to the digit "1", placed 15 times at the beginning and one time at the end of the decoded response. The digit deletions were approximately evenly distributed (31 recurred for the first digit, 29 for the second and 29 for the last). Two of the substitutions recurred for the first digit, five for the second and three for the last.

*Discussion*

Overall, our proposed DIN test setup showed low WER across participants, demonstrating the potential of Kaldi-NL as a mechanism for automating the scoring of the DIN test. These results are likely to be related to the wide variety of recordings used as training material for Kaldi-NL. Nevertheless, there could be a bias for our population, since it included staff members of the Department of Otorhinolaryngology, who could be familiar with speech tests (more specifically, speech audiometry tests), resulting in participants given unusually clear spoken responses (benefitting the ASR).

While the DIN test showed high performance for most participants, Kaldi-NL showed poor performance (48% WER) for participant 27, who spoke both quickly and quietly. No specific instructions were given to the participants regarding how loud, clear and/or at what pace they had to speak, as well as to what words to use (i.e., digit or non-digit words). Therefore, a potential improvement for cases like participant 27 would be to train how to speak or to implement some speaker adaptation in the ASR system (according to specific speaking patterns of a particular participant), a valuable option for users who have recurring audiological evaluations.

To explore potential age effects, we divided our group into two. The t-test results showed no significant difference between the means of the WER for the two groups. Moreover, the ages of the six participants (25, 33, 42, 44, 50 and 54 years old) with which the ASR scored 0% WER indicate there is no evident preference for the participant's age in order for Kaldi-NL to achieve its highest performance.

All 16 insertions corresponded to the digit "1", placed 15 times at the beginning of the decoded spoken response. For 14 of these 15, a segment of noise occurred before the digits. This suggests that the noise was decoded as "1". The overall number of insertions could be decreased by ensuring that the system starts recording the participant's spoken response only after the presentation of the DIN triplet has finished (i.e., avoiding the presence of noise in the recordings).

Based on the given results and the adaptive procedure of the DIN test, the overall effect of the decoding errors from Kaldi-NL on the participants' DIN test SRT; i.e., how much the obtained Kaldi-NL SRT using our proposed setup deviates from a hypothetical SRT calculated without decoding errors by, for example, a clinician, remained unclear. Therefore, our next step was to simulate the introduction of decoding errors.

**Study 2: Evaluation of the automated DIN Test SRT**

*Methods*

The purpose of Study 2 was to analyse the effect of the introduction of decoding errors from Kaldi-NL on the DIN test SRT. To achieve this, the DIN test results for the six participants from Study 1, for whom Kaldi-NL showed 0% WER, were used to model what number of triplets with decoding errors would be acceptable for real-life applications.

*Procedure*

We used the bootstrapping method to run simulated DIN tests in MATLAB (MATLAB, 2019). For each of the six cases, the first step was to list all the SNRs at which the triplets were presented during the DIN test, and to compute the participant's probability to score a correct response (i.e., where all spoken and recognised digits matched the presented triplet) or incorrect response (i.e., one or more spoken and recognised digits did not match those in the presented triplet) to stimuli presented at each of the aforementioned SNRs, creating a sampling frame per participant.

The next step was to simulate 10000 DIN test cases for each of the six participants. Every simulated response was calculated by pseudorandomly drawing scores from the sampling frame. A Gaussian distribution was then fitted to the simulated data per participant, estimating its mean ($\mu_{SRT_{sim}}$) and standard deviation ($\sigma_{SRT_{sim}}$).

The third step was to simulate 10000 times the effect of the inclusion of 1 to 23 triplets with Kaldi-NL decoding errors, using the same simulation structure. The starting SNR in the simulation was equal to the SNR at which the participant's response to the first triplet was scored as correct (from the original DIN test). The index of the triplet where the decoding error was inserted was pseudorandomised within the range of 2–24, as the DIN test setup required that the response to the first triplet was marked as correct for the subsequent triplets to be presented. The impact of the decoding error on the scoring of the response to its corresponding triplet was also pseudorandomly drawn from the sampling frame of the probability of the effect of the decoding error given a correct or incorrect spoken response (given by equations 1-4). A Gaussian distribution was fitted to the simulated inclusion of decoding error(s), and its mean ($\mu_e$) and standard deviation ($\sigma_e$) were computed for each of the decoding error scenarios. Finally, all 23 estimated means of the distributions with decoding errors were compared to that of the sampling distribution without decoding errors.

*Results*

The means of the modelled SRT sampling distributions for all six selected participants are listed in Table 2. The largest difference between the mean of the original DIN test score ($\mu_{SRT_{Kaldi-NL}}$) and of the simulated sampling distribution before the inclusion ($\mu_{SRT_{sim}}$) of decoding errors is 0.17 dB. As the number of inserted decoding errors increased, both $\mu_e$ and $\sigma_e$ increased, as expected. Since the distributions for all six participants showed the same pattern, the performance of only participant 1 is discussed. Figure 4 shows the $\mu_e$ and $\sigma_e$ with the inclusion of 0–23 decoding errors. Figure 5 illustrates the effect of the inserted decoding errors on the simulated DIN SRT, showing histograms and Gaussian-fitted distributions of the 24 simulated scenarios (including no decoding error).

**Table 2.** Kaldi-NL DIN scores and estimated means of simulated SRT sampling distributions ($\mu_{SRT_{sim}}$).

| Participant | Original Kaldi-NL DIN test score | | Modelled sampling distributions | Deviation of mean of simulated SRT sampling distribution from original |
|---|---|---|---|---|
| | $\mu_{SRT_{Kaldi-NL}}$ (dB SNR) | $\sigma_{SRT_{Kaldi-NL}}$ (dB SNR) | $\mu_{SRT_{sim}}$ (dB SNR) | $|\mu_{SRT_{Kaldi-NL}} - \mu_{SRT_{sim}}|$ (dB) |
| 1 | -7.3 | 1.4 | -7.3 | 0.06 |
| 8 | -7.3 | 1.9 | -7.1 | 0.17 |
| 15 | -7.4 | 1.4 | -7.3 | 0.13 |
| 16 | -7.7 | 1.5 | -7.5 | 0.13 |
| 19 | -9.0 | 1.4 | -9.1 | 0.09 |
| 21 | -7.9 | 1.4 | -7.9 | 0.01 |

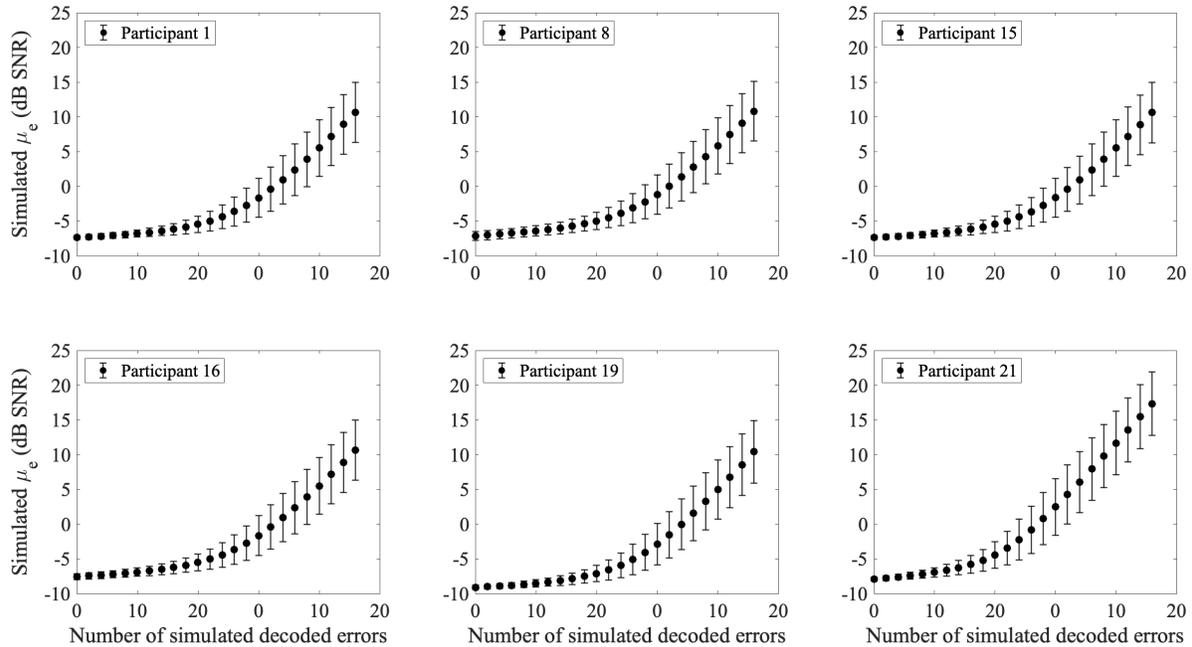

**Figure 4.** Means and standard deviations of the Gaussian distributions for the simulated inclusion of 0-23 triplets with decoding errors for six participants.

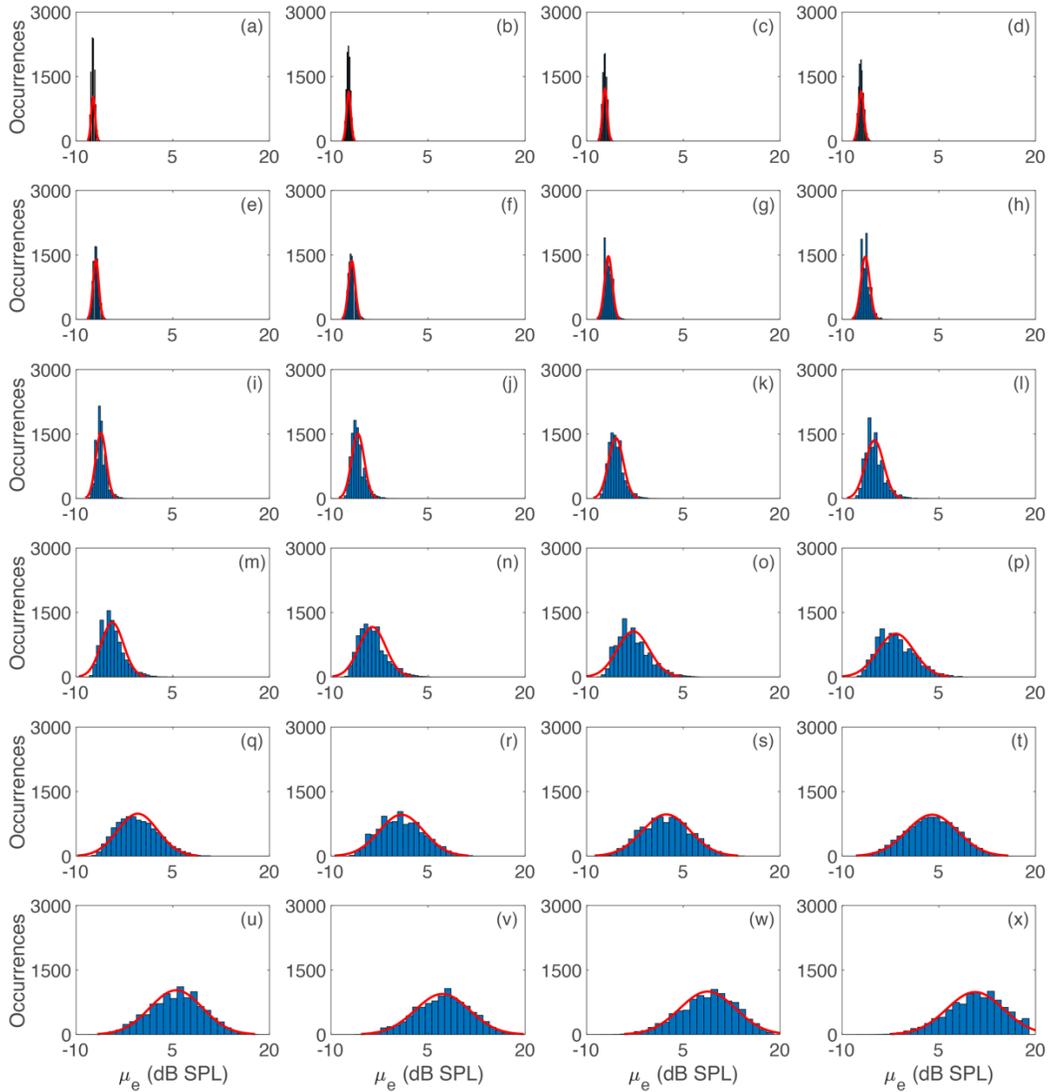

**Figure 5**. Histograms and Gaussian-fitted distributions of the simulated sampling distributions with the inclusion of 0–23 triplets with decoding errors for participant 1. Panel (a) is for zero decoding errors. Panels (b) to (x) show the progressive change of the simulated sampling distributions as the number of inserted decoding errors increases. Histograms are depicted in blue and the Gaussian-fitted distributions in red.

*Discussion*

The modelling of the sampling distributions with the inclusion of decoding errors showed a consistent pattern across participants. The values of $\mu_e$ and $\sigma_e$ of the sampling distributions increased as the number of decoding errors increased, as expected. With up to four triplets with decoding errors per participant, the proposed DIN test setup still provided a clinically reliable result for normal-hearing native Dutch speaking adults, as shown in Table 3, based on the comparison increment of $\mu_e$ for all six participants to the documented 0.70 dB within-subject variability for young normal-hearing adults (Smits et al., 2013). Up to four modelled triplets with decoding errors resulted in a deviation from the original DIN test score ($\mu_{SRT_{Kaldi-NL}}$) of ≤ 0.71 dB, as shown in the rightmost column of Table 3. Depending on the SRT evaluation requirements (e.g., how precise the result should be), clinicians and/or researchers could refer to Figure 5 to define what number of triplets with decoding errors using our DIN test setup would be acceptable for a particular case.

**Table 3.** Kaldi-NL DIN scores and estimated means of simulated SRT sampling distributions with four triplets with decoding errors ($\mu_e$, where $e = 4$). The rightmost column shows the deviation of the estimated mean of the simulated SRT sampling distributions with four triplets with decoding errors ($\mu_e$, where $e = 4$) from the original DIN score ($\mu_{SRT_{Kaldi-NL}}$).

| Participant | Original Kaldi-NL DIN test score $\mu_{SRT_{Kaldi-NL}}$ (dB SNR) | Modelled sampling distributions $\mu_e$, where $e = 4$ (dB SNR) | Deviation of mean of simulated SRT sampling distribution from original $\|\mu_{SRT_{Kaldi-NL}} - \mu_e\|$, where $e = 4$ (dB) |
|---|---|---|---|
| 1 | -7.29 | -6.93 | 0.36 |
| 8 | -7.29 | -6.58 | 0.71 |
| 15 | -7.48 | -6.93 | 0.55 |
| 16 | -7.67 | -7.07 | 0.60 |
| 19 | -9.00 | -8.66 | 0.34 |
| 21 | -7.90 | -7.20 | 0.70 |

**General Discussion**

In recent years, there has been increasing interest in exploring applications combining ASR systems and speech audiometry (Deprez et al., 2013; Meyer et al., 2015; Ooster et al., 2018, 2019, 2020; Pragt et al., 2022; Yilmaz et al., 2014). All these applications involve the training and use of ASR systems to automate the scoring of speech audiometry tests, most often the OLSA test (Meyer et al., 2015; Ooster et al., 2018, 2019, 2020). The present results show that using a simpler and more accessible system, with Kaldi-NL to automatically score the DIN test can produce reliable results.

Study 1 showed that on average there were three triplets with decoding errors from Kaldi-NL per participant. Based on Study 2, this translates to a deviation from the DIN test SRT (without decoding errors from the ASR) less than the within-subject variability (0.70 dB) for DIN test scores of young normal-hearing adults (Smits et al., 2004), as shown by the modelled four triplets with decoded errors in Table 3. These findings are in line with those of Ooster et al. (2018), whose test-retest standard deviations using the SRT obtained through their ASR and through the human supervisor were comparable to the variances in literature for normal-hearing participants (in the range of 0.5 dB). Similarly, the observed effects of the inserted decoding errors from Kaldi-NL on the modelled DIN test SRTs (Study 2), match those described in Pragt et al. (2022) and Yilmaz et al. (2014). This demonstrates that the design of the DIN test (step size for adaptivity, number of triplets and scoring method) is able to reduce or counteract the effect of errors in up to four triplets derived from Kaldi-NL or from the wrong classification of digits by a human supervisor.

The main advantages of the use of Kaldi-NL with the DIN test are the relative simplicity of its implementation and its accessibility. While it might require specific expertise to implement, compared to many other ASRs Kaldi-NL does not require specific training, retraining or further development. It is widely available as an open-source system, does not require high computational power to run and, as observed, offers results with low WER for normal-hearing Dutch native speaking adults. The main limitations of the proposed setup are: 1) the decrease in accuracy when the quality of the audio recordings (more specifically the volume of the recorded speech and potential presence of external noise, such as room acoustics) decreases, 2) the speech material used to train the ASR (which may or may not include different Dutch accents), and 3) that Kaldi-NL is optimised to decode speech using contextual information of the spoken material (i.e., context of a conversation) relying on its language model (which can be challenging as the recorded speech to decode for the DIN test involves only a few digits without more contextual information).

The remaining questions are whether our results could generalise 1) to other populations and 2) to other speech audiometry tests. Nevertheless, new challenges including an increased WER are to be expected in both cases. When testing other populations such as non-native Dutch speaking adults and children (both normal-hearing and hearing-impaired), specific tailoring of Kaldi-NL might be required [as proposed in Ooster et al. (2023)]. This is related to the fact that ASR systems are usually trained using speech material from healthy young to middle-aged adults (Hämäläinen et al., 2015), and that speech varies markedly within and between other populations (Benzeghiba et al., 2007; Bigham et al., 2017; Glasser, 2019). Examples of these variations include different pronunciations, speaking rates, long-term modulations of the voice (e.g., emphasising or denoting emotions) and changes in the quality of the voice due to physiological or behavioural factors (Benzeghiba et al., 2007). Physiological reasons are mostly related to changes in the larynx. Examples of larynx changes include growing in length and thickness as children develop (Kahane, 1982), as well as cartilage calcification, reduced flexibility and muscle atrophy as adults age (Martins et al., 2015). Larynx changes may alter the voice characteristics and speaking styles, and have been the focus of multiple studies looking for improvements in ASR systems by, for instance, suggesting the normalisation of the speaker's vocal-tract length and/or pitch (Gerosa et al., 2009; Liao et al., 2015; Shahnawazuddin et al., 2017), which are two speaker-specific voice cues. These normalisations are usually implemented in ASR systems trained on adult's speech to help alleviate the impact of different vocal-tract lengths and/or pitches among speakers; e.g., children, resulting in the ASR becoming less sensitive to these variations. Another example of suggested improvements regards the inclusion of speech produced by specific populations for the training of the ASR, resulting in an enhanced recognition accuracy by allowing the system to adapt to characteristics specific to these populations; e.g., pronunciation, vocabulary, intonation, speaking rate, vocal-tract length and pitch (Benzeghiba et al., 2007; Gerosa et al., 2009; Glasser, 2019; Hämäläinen et al., 2015).

When using Kaldi-NL for the automatic scoring of speech audiometry tests where the speech material does not have much contextual information (perhaps even less than the DIN test), it is also expected that the ASR performance will drop significantly. The main reason for this is that most ASR systems have a language model that predicts the probability of a sequence of words in a sentence; thus, relying on context. One particular example of a speech audiometry test with no contextual information is the consonant-vowel-consonant (CVC, three-phoneme) word lists, widely used in clinical audiological evaluation (especially in the Netherlands), and scored based on individual phoneme recognition. A possible way to overcome this problem for this type of test can include replacing the existing language model for a character-level one (which predicts the probability of a sequence of individual characters; i.e., symbols in written language, rather than words) (Al-Rfou et al., 2019; Hua et al., 2018).

Another potential solution for both testing other populations or automating other speech audiometry tests could be using a sequence-to-sequence model. Sequence-to-sequence models have been gaining popularity by using a single neural network instead of separate acoustic, pronunciation and language models (Chiu et al., 2018). An example of a promising robust sequence-to-sequence model is Whisper, an architecture that has shown high-quality results without the need of fine tuning any dataset (Radford et al., 2023). Depending on the new challenges we may face when testing our proposed system with different populations and other speech audiometry tests, we shall explore some of these suggestions from literature in the near future.

**Conclusions**

We have presented a simple system for automation of the scoring of the DIN test using the widely available ASR for the Dutch community, Kaldi-NL. When testing native Dutch speaking normal-hearing adults, the estimated DIN test SRTs from Study 1 suggest that the proposed system gives comparable results to the currently used clinical setup (where a human manually scores the DIN test). Study 2 showed, through the simulation of Kaldi-NL decoding errors, that up to four triplets with decoding errors do not produce a variation larger than 0.70 dB.

**Acknowledgements**

The authors would like to thank ir. Bert Maat from the University Medical Center Groningen, for their suggestions and support throughout the study, and Prof. dr. ir. Cas Smits from Amsterdam University Medical Center, for sharing the DIN test stimuli.

**Funding**

This work was funded by VICI grant 918-17-603 from the Netherlands Organization for Scientific Research (NWO) and the Netherlands Organization for Health Research and Development (ZonMw). Further support was provided by the Heinsius Houbolt Foundation and the W.J. Kolff Institute for Biomedical Engineering and Materials Science from University Medical Center Groningen, University of Groningen. The study is part of the research program of the UMCG Otorhinolaryngology Department: Healthy Aging and Communication.

**Notes**

Data for this study will be publicly accessible in DataverseNL.